\documentclass[letterpaper, 10 pt, conference]{ieeeconf}  

\IEEEoverridecommandlockouts                              

\usepackage{amsmath} 
\usepackage{amssymb}  
\usepackage{multirow}
\usepackage{xspace}
\usepackage[linesnumbered,ruled,vlined]{algorithm2e}
\usepackage{booktabs}
\usepackage[capitalize]{cleveref}
\usepackage{url}
\usepackage{todonotes}
\usepackage{graphicx}
\usepackage{pifont}
\newcommand{\cmark}{\ding{51}}%
\newcommand{\xmark}{\ding{55}}%
\newcommand{\squeezesmall}{\vspace{-1mm}}%
\makeatletter
\DeclareRobustCommand\onedot{\futurelet\@let@token\@onedot}
\def\@onedot{\ifx\@let@token.\else.\null\fi\xspace}

\def\eg{\emph{e.g}\onedot} 
 
\def\cf{\emph{cf}\onedot}

\def\etal{\emph{et al}\onedot}
\makeatother

\title{\LARGE \bf
X2Graph for Cancer Subtyping Prediction on Biological Tabular Data
}

\author{Tu Bui$^{1,*}$ and Mohamed Suliman$^{1}$ and Aparajita Haldar$^{1}$ and Mohammed Amer$^{1}$ and Serban Georgescu$^{1}$
\thanks{$^{1}$Tu Bui, Mohamed Suliman, Aparajita Haldar, Mohammed Amer and Serban Georgescu are with Fujitsu Research of Europe,
        Slough, The United Kingdom.}%
\thanks{{*\tt\small tu.bui@fujitsu.com}}%
\thanks{© 20XX IEEE.  Personal use of this material is permitted.  Permission from IEEE must be obtained for all other uses, in any current or future media, including reprinting/republishing this material for advertising or promotional purposes, creating new collective works, for resale or redistribution to servers or lists, or reuse of any copyrighted component of this work in other works.}%
}

\begin{document}

\maketitle
\thispagestyle{empty}
\pagestyle{empty}

\begin{abstract}
Despite the transformative impact of deep learning on text, audio, and image datasets, its dominance in tabular data, especially in the medical domain where data are often scarce, remains less clear. In this paper, we propose X2Graph, a novel deep learning method that achieves strong performance on small biological tabular datasets. X2Graph leverages external knowledge about  the relationships between table columns, such as gene interactions, to convert each sample into a graph structure.  This transformation enables the application of standard message passing algorithms for graph modeling. Our X2Graph method demonstrates superior performance compared to existing tree-based and deep learning methods across three cancer subtyping datasets. 
\newline

\indent \textit{Clinical relevance}— This work advances the application of deep learning solutions to cancer diagnosis, particularly in scenarios where only limited tabular data is available.
\end{abstract}

\section{INTRODUCTION}
\label{sec:intro}

The rise of computer-assisted systems in medical diagnosis and prognosis marks a transformative period in healthcare, driven by rapid advancements in artificial intelligence (AI) and Deep Learning (DL). These technologies have ushered in a new era of precision and efficiency, enabling healthcare professionals to leverage vast amounts of patient data to deliver highly accurate and personalized diagnoses in a timely manner. However, most recent medical AI/DL applications are centered around only the image domain, according to the World Economic Forum 2025 findings \cite{wef25}. These unbalanced developments are also observed in other sectors besides healthcare -- tremendous progress has been acquired from deep learning on image and text, however on other domains such as tabular data, statistical and tree-based methods still remain dominant \cite{grinsztajn2022tree}. The main reason for the lack of a wide adaptation of DL on tabular data comes from the inherent challenges of tabular data itself. Breugel \etal \cite{breugel24iclr} highlights some of these challenges: (i) tabular dataset often contains mixed type data \eg continuous, categorical, date time; (ii) is moderate in size; (iii) cell values are related to contextual metadata \eg column names; and (iv) is invariant/equivariant with respect to the column order, encouraging the modeling methods to also have such invariance. 

Within the health care sector, there are plentiful examples of biological tabular data (BioTD), from clinical records to genetic data such as copy number variation (CNV) and RNA expression. Clinical data, including patient demographics, lab results and medical history, provides an essential context that enriches the interpretation of imaging and molecular data, enabling AI models to make better predictions \cite{esteva2019guide}. CNV data offers valuable insights into genetic alterations often associated with diseases, and its inclusion in AI models facilitates a more comprehensive understanding of genetic contributions to health conditions \cite{almal2012implications}. RNA sequencing data further augments the predictive power of the model by revealing the transcriptomic landscape, which reflects the functional state of genes and can uncover disease mechanisms that may not be apparent through imaging alone \cite{eswaran2012transcriptomic, huang2011rna}. Collectively, these diverse data modalities empower AI models not only to predict outcomes with greater precision but also to uncover novel insights into disease mechanisms \cite{esteva2019guide, jiang2017artificial}. Again, these modalities pose even greater challenges in modeling tabular data with DL. Medical data are scarce due to their sensitive nature and strict requirements for patient privacy preservation. The high time and labor cost of experiments also limit the amount of annotated data. On the other hand, CNV and RNA data tend to have a vast number of genes, which significantly reduces the samples-to-features ratio, and therefore, worsens the risk of overfitting in training DL models. The development of DL methods specifically tailored for BioTD is an urgent priority.

We propose X2Graph, a novel DL method to model BioTD. X2Graph addresses the aforementioned challenges of tabular data by converting each row sample into a view-invariant graph and modeling these graphs with standard message-passing neural networks. X2Graph leverages external domain knowledge abundant in literature to define the node connections, therefore strengthening the inductive bias during model learning and reducing overfitting. Our X2Graph contributions are threefold:
\begin{itemize}
    \item A method to convert BioTD samples to graphs and model them with graph neural networks, in contrast with existing DL methods mostly based on multilayer perceptrons (MLP) or attention mechanism.
    \item X2Graph achieves state-of-art performance on three real-life BioTD datasets, covering three distinct modalities (CNV, RNA, Clinical) when compared with various DL and statistical methods.
    \item A graph-based interpretability study into X2Graph models, highlighting the key contributing components and important features that underpin its predictions. 
\end{itemize}

\begin{figure*}[thpb]
    \centering
    \includegraphics[width=0.70\linewidth,trim=0cm 0cm 0cm 0cm,clip]{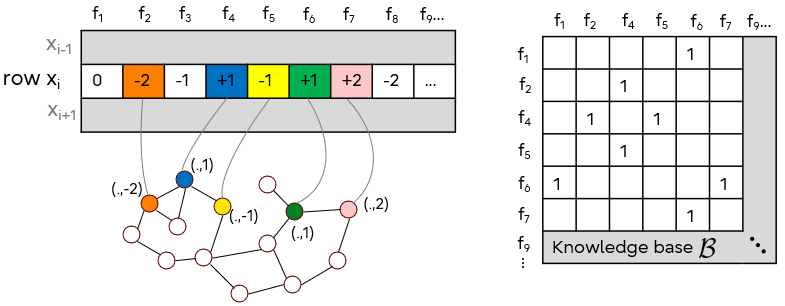}
    \caption{X2Graph converts each table row into a graph. The cell values become node features, while the edge connection comes from the KB. The (.) notation refers to other information that may be incorporated alongside the cell value, for example, the feature name or feature index. Note: not all cells at row $x_i$ may appear on the graph as in the case of $f_3$ and $f_8$ because they are not available in the KB above; also, $f_1$ is also dropped according to certain assumptions, e.g. value of 0 is not meaningful for the modeling task. }
    \label{fig:x2graph}
\end{figure*}

\section{BACKGROUND}







Early approaches for modeling tabular data are based on feature selection \cite{lasso}, regression \cite{marquardt1975ridge} or rule-based shallow methods such as decision trees and random forest \cite{breiman2001random}. XGBoost \cite{chen2016xgboost} further improves accuracy and robustness by sequentially building trees that correct the errors of the previous ones. Being also a computationally efficient and interpretable algorithm, XGBoost is still a preferred choice for modeling small to medium-tabular data to date \cite{grinsztajn2022tree}.

By contrast, DL methods can handle complex, high-dimensional relationships and scale to large datasets but are less interpretable and often require massive training data. In the healthcare domain, MLP-based methods are widely used on tabular data, often bootstrapped with an image model (whose training data is more abundant) in a multimodal fashion \cite{wang2021lung,liu2022hybrid,amer}.  
Transformer-based models, meanwhile, leverage attention mechanisms to capture complex dependencies between features \cite{hollmann2022tabpfn,arik2021tabnet,huang2020tabtransformer} at the cost of computational complexity. 
Challenges such as overfitting and spurious correlations are persistent issues that are exacerbated by insufficient training data \cite{shwartz2022tabular, liu2017deep, ye2024spurious}. Self-supervised learning (SSL) could alleviate these challenges \cite{yoon2020vime}, but requires a large amount of extra data, often unlabeled and in different distributions, which in turn, adds complexity to the learning task.   
\begin{figure*}[thpb]
    \centering
    \includegraphics[width=0.8\linewidth,trim=0cm 0cm 0cm 0cm,clip]{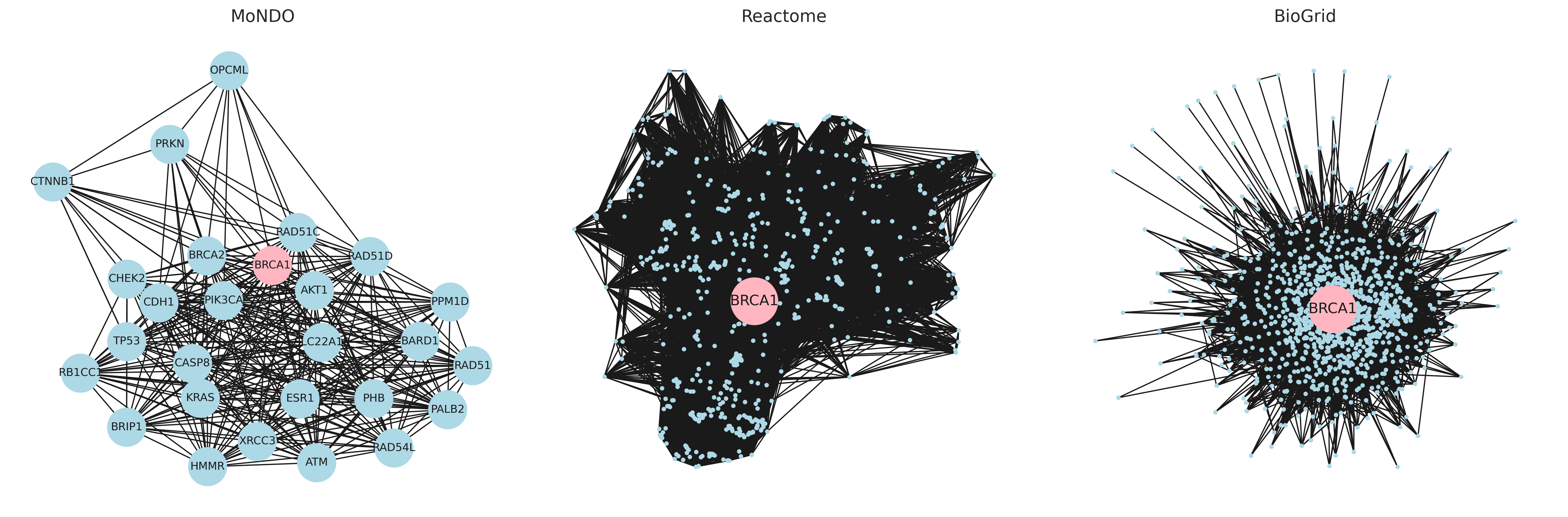}
    \caption{Subgraph visualization for three gene KBs. Each subgraph shows 1-hop neighbor connections centered at gene BRCA1.}
    \label{fig:kb}
\end{figure*}

Graph DL models, such as Graph Neural Networks (GNNs), are becoming increasingly popular due to their strength in capturing complex relationships and inter-dependencies between features in tabular data. Unlike MLP methods in which all neurons are connected, GNN performs message passing within nodes connected by an edge only. Overfitting is, therefore, constrained through the inductive bias defined by the graph edges. A common method to model tabular data with GNNs is to convert the whole table into a large graph whose nodes correspond to either rows (samples), columns (features) or both (heterogeneous), while edges represent node similarity according to a predefined distance metric, as described in the taxonomy paper GNN4TDL \cite{li2023graph}. IGGNet \cite{iggnet} converts each row into a graph, however inductive bias is still obtained intrinsically from the data itself through feature linear correlation, which is susceptible to noises and outliers. Likewise, TABGLM \cite{tabglm} treats each row as a fully connected structure graphs and combines with a language model for table header semantic modeling; yet it is unclear if their method generalizes to datasets with low samples-to-features ratio.  In contrast, our proposed method leverages external knowledge to define the graph edges, offering a more accurate inductive bias for model learning. The abundant external knowledge (henceforth knowledge base, or KB) in the biological domain is also advantageous for X2Graph. The closest work to ours is Plato \cite{ruiz2024high}, which also employs a KB for edge definition. However, Plato requires a KB much larger than the tabular dataset so that an embedding for each feature could be learned effectively via SSL. This does not apply to our problem settings where the KB contains just a subset of features of the tabular datasets. 

\begin{algorithm}[t]
\DontPrintSemicolon
  \newcommand\mycommfont[1]{\footnotesize\ttfamily\textcolor{blue}{#1}}
  \SetCommentSty{mycommfont}
  \SetKwInput{KwInput}{Input}                
  \SetKwInput{KwOutput}{Output}              
  \KwInput{Table $\mathbf{\mathcal{D}}=\{\mathbf{S}_D, \mathbf{X}_D\}$ with N rows and D columns, $\mathbf{S}_D=\{f_1, f_2,...,f_D\}$ list of column names; $\mathbf{X}_D\in \mathbb{R}^{N\times D}$ the value matrix. }
  \KwData{Knowledge base $\mathbf{\mathcal{B}}=\{\mathbf{S}_B, \mathbf{X}_B\}$ where $\mathbf{S}_B$ is the list of B features, $\mathbf{X}_B\in \mathbb{R}^{B\times B}$ is the square connection matrix.}
  \KwOutput{Graph dataset $\mathbf{\mathcal{G}}=\{\textbf{G}_i\}$, $i=1,2,...,N$}

  $\mathbf{S} \gets \mathbf{S}_D \cap \mathbf{S}_B$ \tcp*{intersect 2 feature lists}
  
  \For{$i\gets 1$ \KwTo $N$}{
    \tcp{Initialize feature list, node vectors and edge indices}
    $\mathbf{\bar{S}}, \mathbf{V}, \mathbf{E} \gets [], [], []$  

    
    \tcp{Construct node vectors}
    \ForEach{$\mathbf{s} \in \mathbf{S}$}{
      $v \gets \mathbf{X}_D(i, \mathbf{s})$
      
      \If{\text{isNodeEligible}($v$)}{  
      
        $\mathbf{\bar{S}}.\text{append}(\mathbf{s})$
        $\mathbf{v}_x \gets \text{nodeFeat}(v, \mathbf{s}, \text{**kwargs})$  

        $\mathbf{V}.\text{append}(\mathbf{v}_x)$
      }
    }

    \tcp{Construct edge indices}
    \ForEach{$\mathbf{s}_1 \in \mathbf{\bar{S}}$}{
      \ForEach{$\mathbf{s}_2 \in \mathbf{\bar{S}}\setminus \mathbf{s}_1$}{
        \If(\tcp*[f]{edge exists in KB}){$\mathbf{X}_B(\mathbf{s}_1, \mathbf{s}_2)!=0$}{  
          $\mathbf{E}.\text{append}((\mathbf{s}_1, \mathbf{s}_2))$
          
          
          
        }
      }
    }

    $\textbf{G}_i \gets \{\mathbf{V}, \mathbf{E}\}$
    }

\caption{X2Graph algorithm. Function $\text{isNodeEligible}(v)$ always return True for RNA and Clinical data. For CNV, it returns True if $v$ is non zero, otherwise False.}
\label{alg:x2graph}
\end{algorithm}

\section{BIOLOGICAL TABULAR DATASETS}
\subsection{TCGA BRCA}
\label{sec:brca}
We study Breast Invasive Carcinoma (TCGA BRCA) - a comprehensive collection of genomic and clinical data for breast cancer patients. TCGA BRCA contains 3 BioTDs - Copy Number Variation (CNV), RNA-seq (henceforth RNA) and Electrical Heath Records (Clinical). 
The target classes are 4 highly-unbalanced cancer subtypes: Luminal A ($53.5\%$), Luminal B ($20.6\%$), Her2-enriched ($7.8\%$) and Basal-like ($18.1\%$). Following \cite{amer}, we only keep 977 patients who have all CNV, RNA and Clinical data. While CNV and RNA data are numerical, Clinical data are both numerical (\eg age at diagnosis), ordinal (\eg IHC scores), and categorical (\eg ICD-10 codes). We convert categorical and ordinal features to one-hot encoding and standardize all Clinical data via z-scoring. 

The number of columns in CNV, RNA and Clinical BioTDs are 23286, 20530 and 138, respectively. Unlike existing tabular datasets in \cite{grinsztajn2022tree} with a medium size ($N>3000$) and medium-to-small columns ($D<500$), the CNV and the RNA datasets have much smaller $N/D$ ratio ($<5\%$), which illustrates a more realistic scenario of a BioTD and makes our problem more challenging. We also use Clinical data for increased diversity in data type (phenotype versus omics) and values (heterogeneity).




\subsection{Knowledge Base Collection}
\label{sec:kb}
For CNV and RNA, we use 3 popular public data sources -- MoNDO \cite{mungall2017monarch}, Reactome \cite{fabregat2018reactome} and BioGrid \cite{oughtred2019biogrid} -- to construct the KBs. Within each network, nodes represent genes (or their protein products), while edges connect genes to each other based on their associations as reflected in that data source. Specifically, MoNDO connects two genes if genome-wide association studies (GWAS) report them both to be associated with the same disease. Reactome connects genes belonging to the same biological pathway group, while BioGrid relies on physical or genetic association between them. We follow the methodology in \cite{nunez2021multilayer} to process and construct each KB. The snapshots of these KBs are visualized in \cref{fig:kb}.

For clinical data, there is not a complete KB about feature relation. Therefore, we manually construct a KB by utilizing previously published research, medical textbooks, and other verified medical resources. In our KB, nodes signify clinical features, while edges connect these features if there is evidence in the literature that they influence each other. For example, feature `HER2 IHC score' (human epidermal growth factor receptor 2 Immunohistochemistry score) and `HER2 FISH status' (HER2 fluorescence in situ hybridization) are strongly correlated according to \cite{furrer2017concordance}. While our clinical KB is not exhaustive, as some nodes and edges may be missing or are not evidence, it poses a practical challenge for our proposed method. 

Overall, the constructed MoNDO KB has 4152 nodes and 54690 edges, excluding self-loops. The numbers of nodes/edges for the Reactome, Biogrid and Clinical KBs are 11716/1937505, 18763/932080 and 50/224, respectively. 

\section{METHODS}
\squeezesmall
Given a BioTD $\mathbf{\mathcal{D}}$ with N rows and D columns, together with targets $Y_D \in \mathbb{R}^N$, we would like to build a machine learning model $f_{\theta}: \mathbf{x_i} \mapsto y_i$ where sample $\mathbf{x}_i$ is row i-th in $\mathbf{\mathcal{D}}$. X2Graph is based on our assumptions that -- if two features are related according to a certain KB, their corresponding nodes/neurons should be connected during the modeling process. We first describe our proposed X2Graph method to convert $\mathbf{\mathcal{D}}$ to a set of graphs $\mathbf{\mathcal{G}}$ in \cref{sec:x2graph}. \cref{sec:graph_model} depicts how $\mathbf{\mathcal{G}}$ is modeled using standard graph neural networks. Finally, \cref{sec:graph_fusion} describes model fusion for a boosted performance. 

\begin{table*}[thpb]
\caption{Details of model architecture for each datasets and knowledge base.}
\label{tab:arch}
\centering
\begin{tabular}{l|ccc|ccc|c}
\toprule
               & \multicolumn{3}{c}{CNV}    & \multicolumn{3}{c}{RNA}        & \multirow{2}{*}{Clinical} \\
               & MoNDO & Reactome & Biogrid & MoNDO    & Reactome & Biogrid  &                           \\
\midrule
Feature dimension ($d$) & 128   & 128      & 512     & 512      & 1024     & 256      & 32                          \\
Architecture   & SG    & GEN      & GCN     & GCN      & SAGE     & GAT      & PNA                       \\
Number of layers    & 2     & 2        & 4       & 2        & 2        & 2        &  3                        \\
Activation     & GELU  & PReLU    & GELU    & GELU     & ReLU     & GELU     & ReLU                      \\
Normalization  & -     & Layer    & Batch   & Instance & Instance & Instance & Batch  \\           \bottomrule       
\end{tabular}
\end{table*}

\begin{figure*}[thpb]
    \centering
    \begin{tabular}{ccc}
      \includegraphics[width=0.3\linewidth]{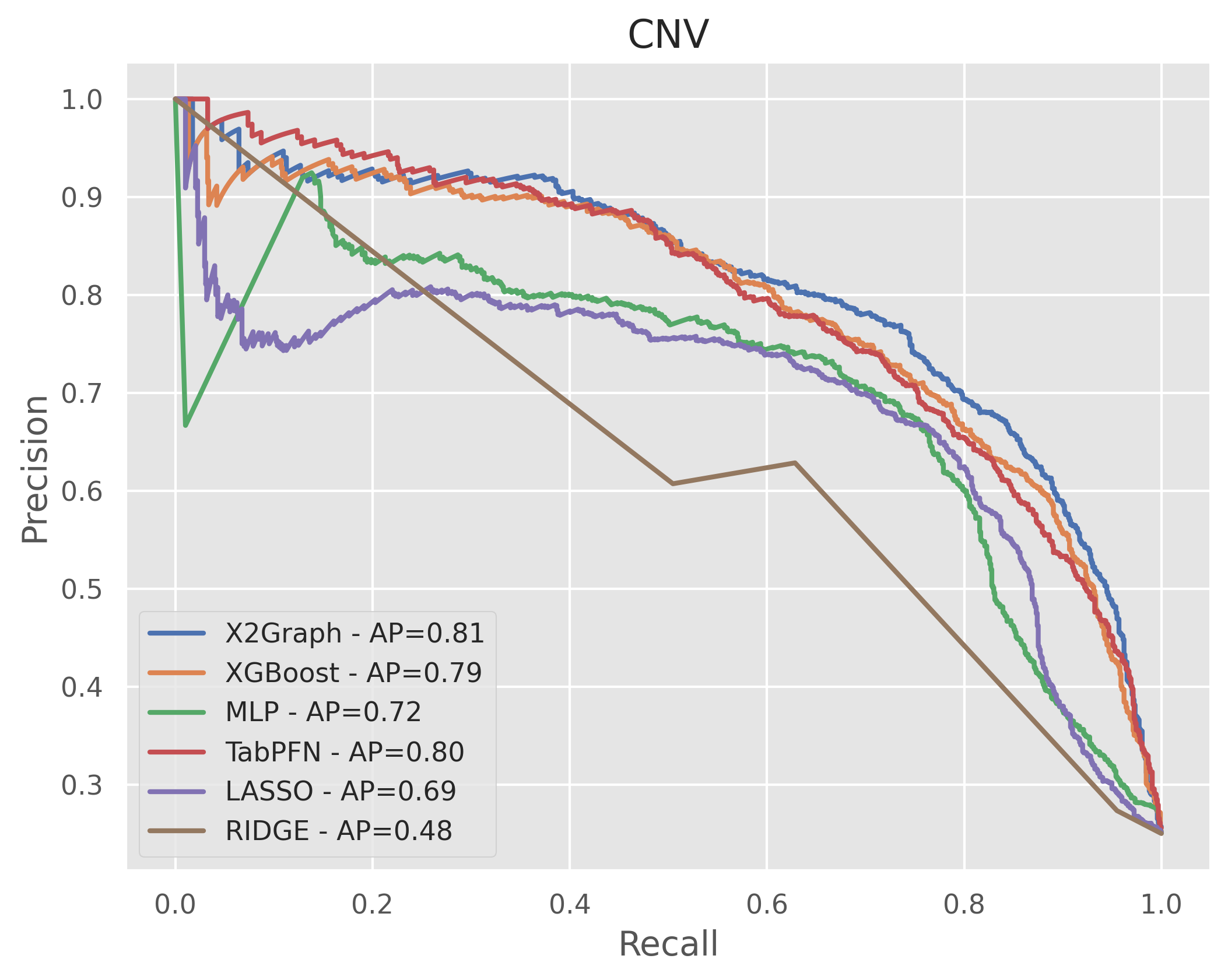}   &  
      \includegraphics[width=0.3\linewidth]{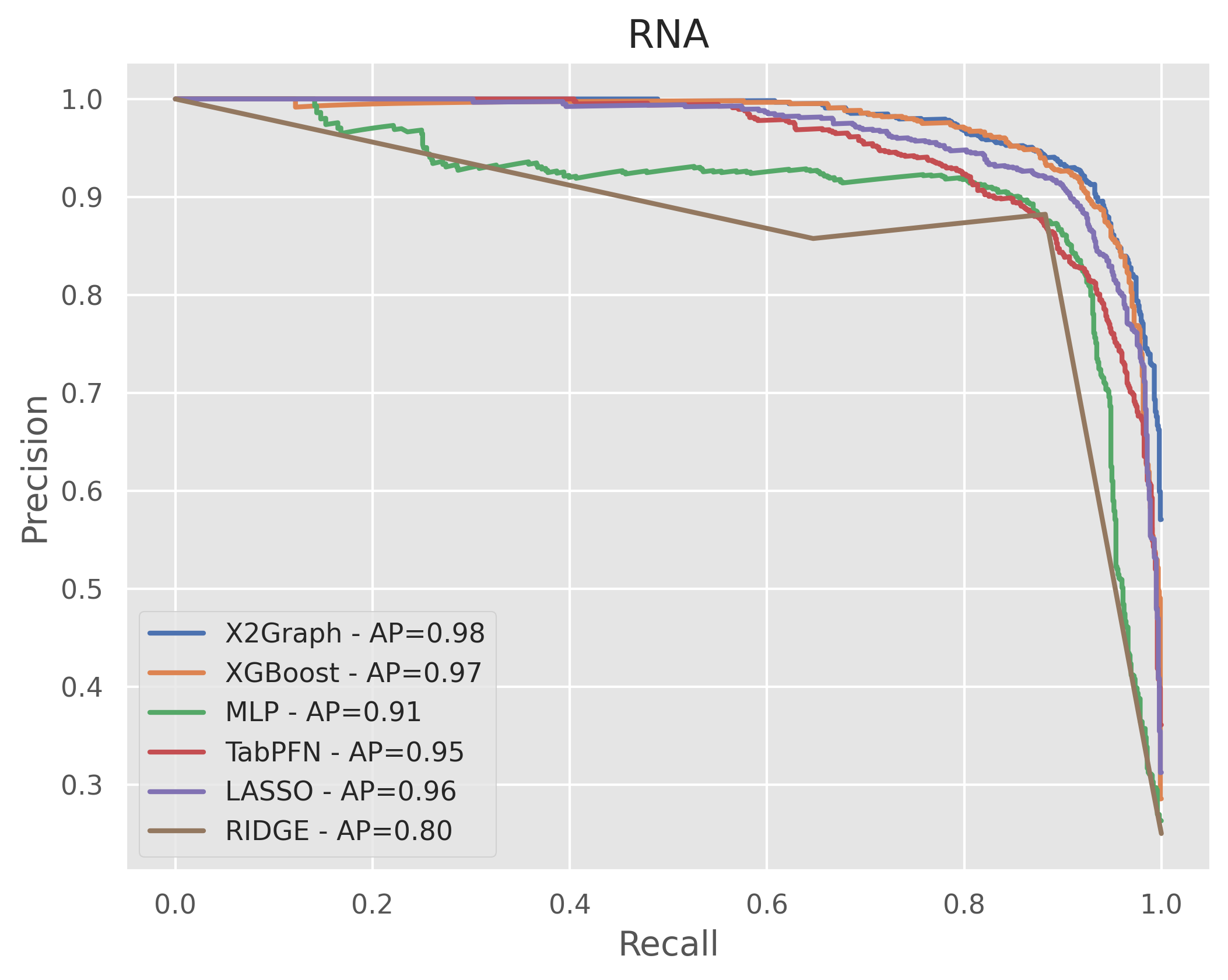} &
      \includegraphics[width=0.3\linewidth]{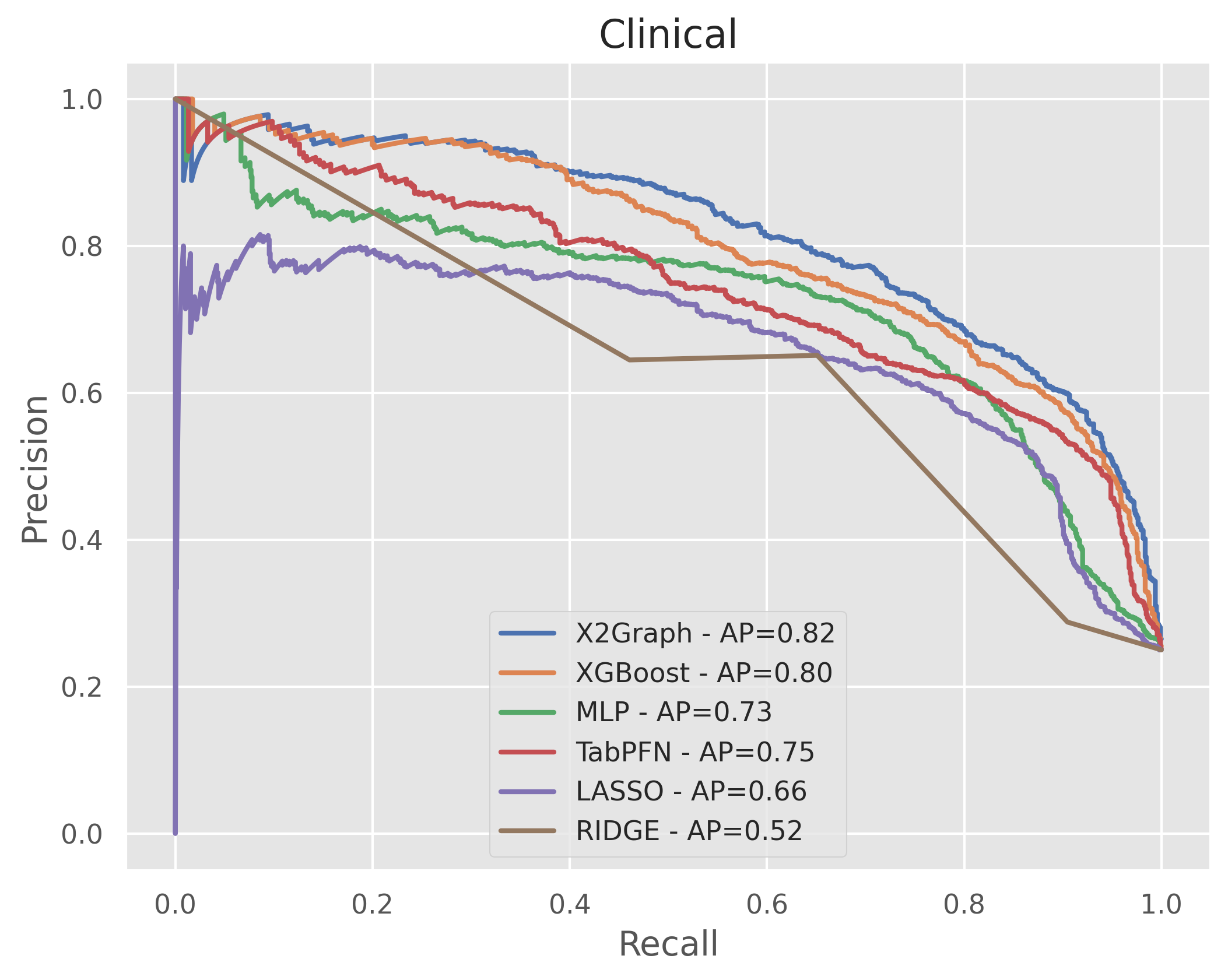}
    \end{tabular}
    \caption{(Top) PR curves and Average Precision (AP) of X2Graph and baselines on the three benchmarks across the 10-fold cross-validation test sets. Note: The curves for RIDGE are different from the rest because RIDGE is a feature selection method which only outputs the predicted class instead of probabilities for each class.}
    \label{fig:pr}
\end{figure*}

\begin{figure*}[thpb]
    \centering
    \begin{tabular}{cccc}
      \includegraphics[width=0.2\linewidth]{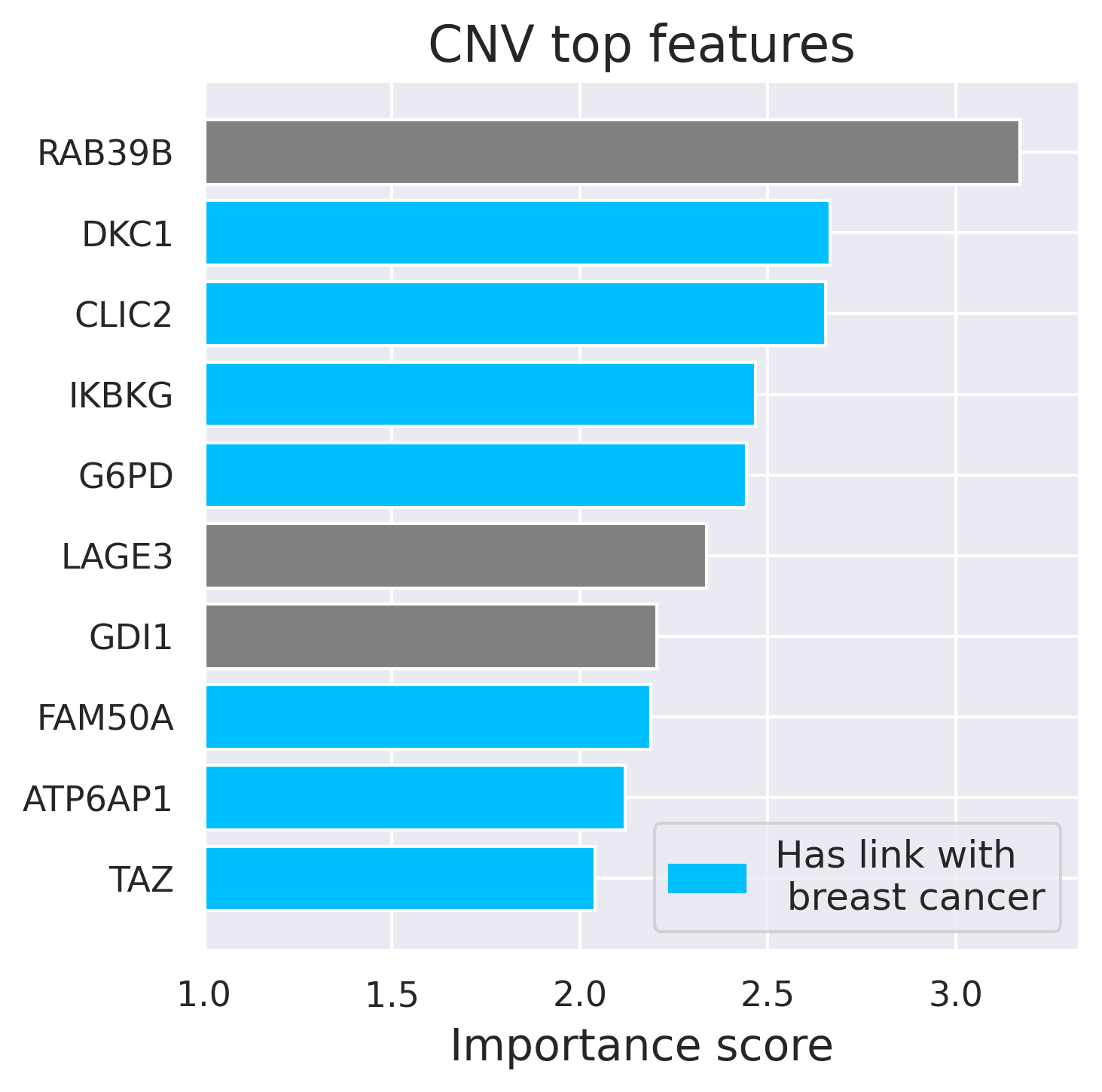}   &  
      \includegraphics[width=0.2\linewidth]{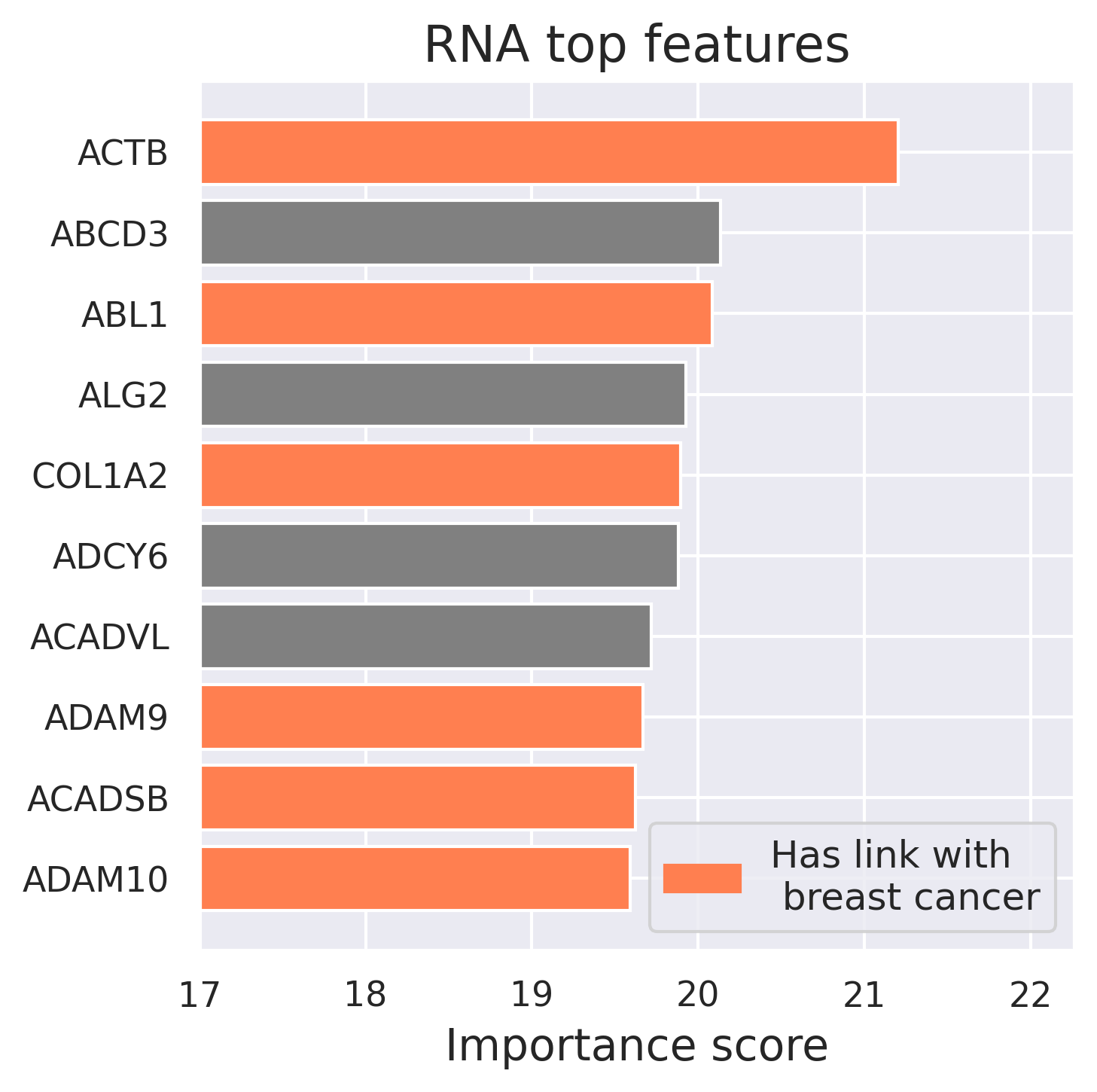} &
      \includegraphics[width=0.27\linewidth]{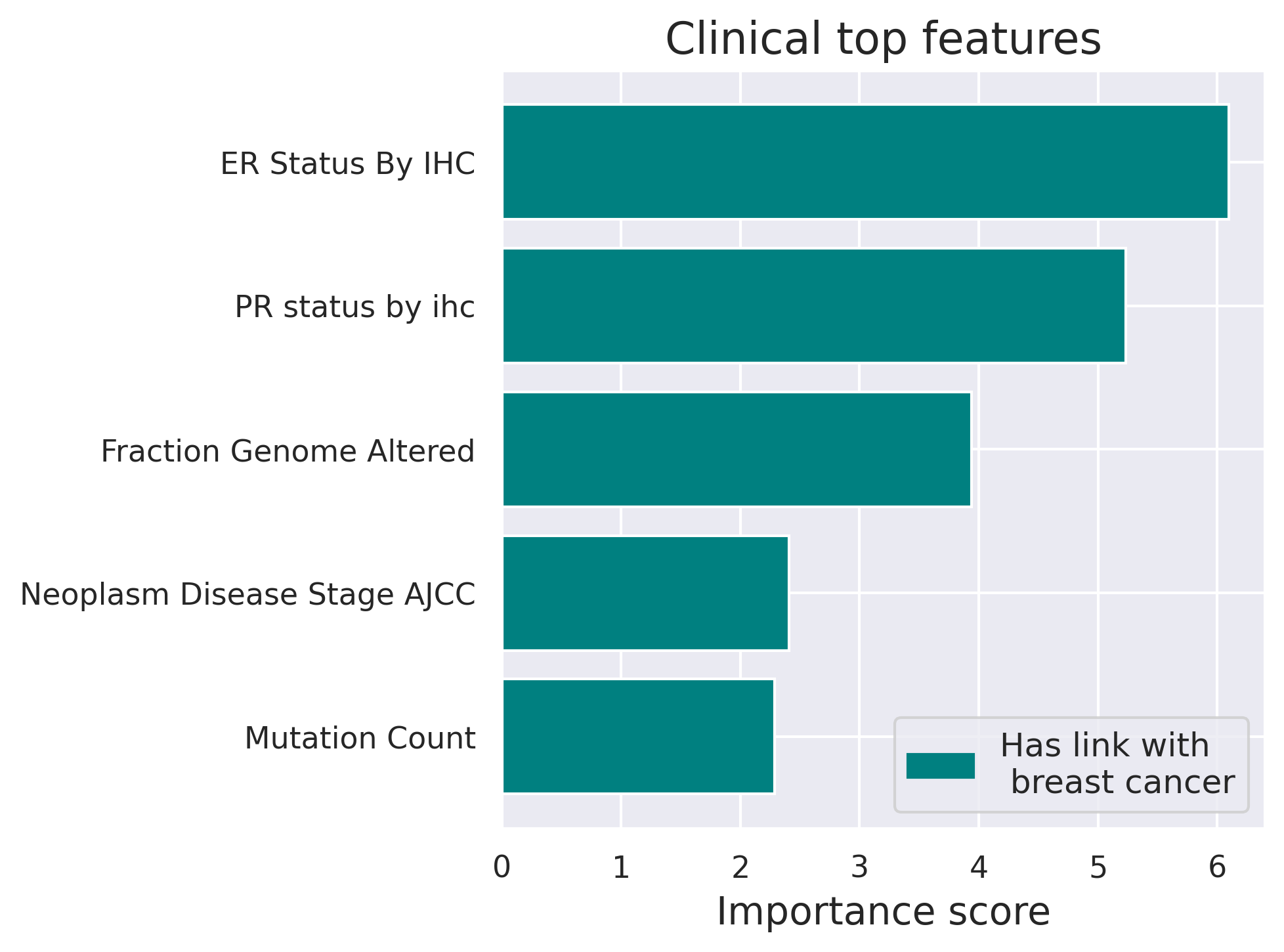} &
      \includegraphics[width=0.2\linewidth]{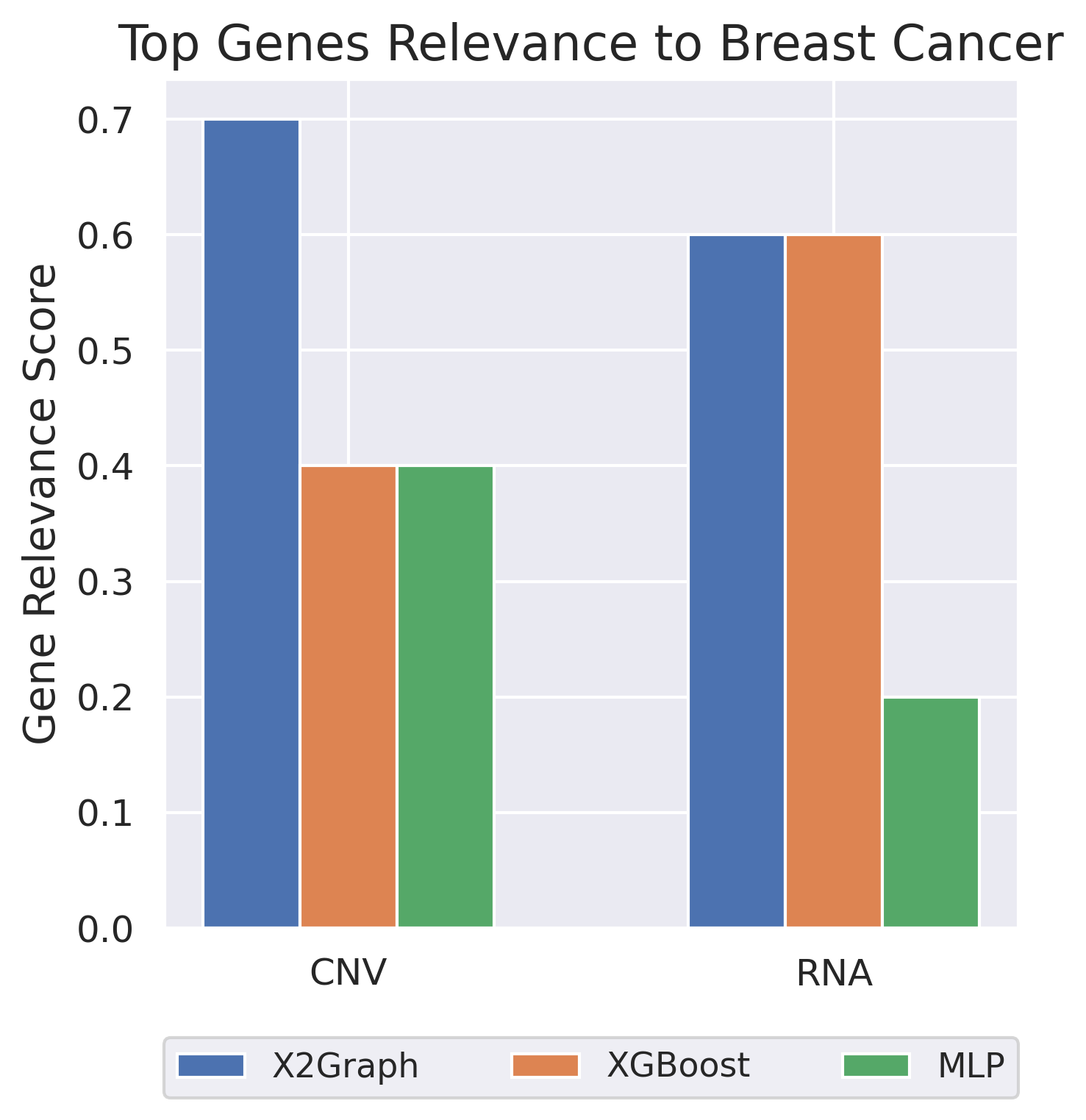} \\
      (a) CNV & (b) RNA & (c) Clinical & (d) Baselines comparison
    \end{tabular}
    \caption{(a-c) Top k most importance features identified from X2Graph models for CNV, RNA and Clinical data. (d) Fractions of these features evidenced in literature.}
    \label{fig:xai}
\end{figure*}

\begin{table}[thpb]
    \caption{X2Graph and baselines on the CNV dataset.}
    \label{tab:cnv}
    \centering
    \begin{tabular}{l|cccc}
         \toprule
         Method &  Accuracy & AUC & F1 & $\mathcal{K}_C$ \\
         \midrule
         X2Graph (fusion) & \textbf{0.7410} & \textbf{0.8935} & \textbf{0.6858} & \textbf{0.5892}\\
         X2Graph Reactome  & 0.7321 & 0.8796 & 0.6815 & 0.5776\\
         X2Graph MoNDO & 0.7239 & 0.8892 & 0.6561 & 0.5635\\
         X2Graph BioGrid  & 0.7143 & 0.8855 & 0.6303 & 0.5421\\
         TabPFN \cite{hollmann2022tabpfn}& 0.7213 & 0.8630 & 0.5875 & 0.5414\\
         MLP \cite{amer}& 0.7025 & 0.8284 & 0.6126 & 0.5283\\
         \midrule
         XGBoost \cite{chen2016xgboost}& 0.7312 & 0.8831 & 0.6837 & 0.5738\\
         LASSO \cite{lasso}& 0.6952 & 0.8344 & 0.6391 & 0.5107\\
         Ridge \cite{marquardt1975ridge}& 0.6291 & 0.6720 & 0.5239 & 0.3815\\
         \bottomrule
    \end{tabular}
\end{table}

\begin{table}[thpb]
    \caption{X2Graph and baselines on the RNA dataset.}
    \label{tab:rna}
    \centering
    \begin{tabular}{l|cccc}
         \toprule
         Method &  Accuracy & AUC & F1 & $\mathcal{K}_C$ \\
         \midrule
         X2Graph (fusion) & \textbf{0.9242} & \textbf{0.9877} & \textbf{0.9098} & \textbf{0.8806}\\
         X2Graph Reactome  & 0.8990 & 0.9856 & 0.8796 & 0.8412\\
         X2Graph MoNDO & 0.9014 & 0.9854 & 0.8671 & 0.8430\\
         X2Graph BioGrid  & 0.8995 & 0.9867 & 0.8804 & 0.8439\\
         MLP \cite{amer}& 0.8805 & 0.9496 & 0.8502 & 0.8080\\
         TabPFN \cite{hollmann2022tabpfn}& 0.8769 & 0.9719 & 0.8434 & 0.8034\\
         \midrule
         XGBoost \cite{chen2016xgboost}& 0.9194 & 0.9857 & 0.8975 & 0.8729\\
         LASSO \cite{lasso}& 0.9040& 0.9821 & 0.8742 & 0.8490\\
         Ridge \cite{marquardt1975ridge}& 0.8826 & 0.8954 & 0.8567 & 0.8098\\
         \bottomrule
    \end{tabular}
\end{table}

\begin{table}[thpb]
    \caption{X2Graph and baselines on the Clinical dataset.}
    \label{tab:cli}
    \centering
    \begin{tabular}{l|cccc}
         \toprule
         Method &  Accuracy & AUC & F1 & $\mathcal{K}_C$ \\
         \midrule
         X2Graph & \textbf{0.7487} & \textbf{0.8930} & 0.6523 & \textbf{0.5797}\\
         MLP \cite{amer}& 0.7043 & 0.8522 & 0.6168 & 0.5074 \\
         TabPFN \cite{hollmann2022tabpfn}& 0.6735 & 0.8493 & 0.6073 & 0.4802\\
         \midrule
         XGBoost \cite{chen2016xgboost}& 0.7250 & 0.8884 & \textbf{0.6800} & 0.5711\\
         LASSO \cite{lasso} & 0.6509 & 0.8349 & 0.6039 & 0.4624 \\
         Ridge \cite{marquardt1975ridge}& 0.6522 & 0.7505 & 0.6083 & 0.4728 \\
         \bottomrule
    \end{tabular}
\end{table}
\subsection{X2Graph}
\label{sec:x2graph}
\cref{fig:x2graph} illustrates our X2Graph conversion. The tabular dataset  $\mathbf{\mathcal{D}}$ is characterized by the list of features (column names) $\mathbf{S}_D$ and the value matrix $\mathbf{X}_D \in \mathbb{R}^{N\times D}$. Assumed we have access to the KB $\mathbf{\mathcal{B}}$ characterized by the list of features $\mathbf{S}_B$ and pairwise relation matrix $\mathbf{X}_B\in \mathbb{R}^{B\times B}$ where $\mathbf{X}_B(s_i,s_j)$ represents the relation between two features $s_i, s_j \in \mathbf{S}_B$. $\mathbf{X}_B$ is a binary square matrix for every KB described in \cref{sec:kb}. We first define the list of features common in both the dataset $\mathbf{\mathcal{D}}$ and KB $\mathbf{\mathcal{B}}$, $\mathbf{S} = \mathbf{S}_D \cap \mathbf{S}_B$, and drop the rest. Next, we convert each row $\mathbf{x}_i$ to a graph $\textbf{G}_i = \{\mathbf{V}, \mathbf{E}\}$ where $\mathbf{V}$ is the vertices and $\mathbf{E}$ is the edges. The edge matrix $\mathbf{E}$ can be constructed directly from the KB:
\begin{equation}
    E=\{(s_u, s_v)\} \forall s_u, s_v \in \mathbf{S} \text{ if } \mathbf{X}_B(s_u, s_v)!=0 \text{ and } u!=v
\end{equation}
The direction of the edges depend on the nature of feature relation in the KB. For those listed in \cref{sec:kb}, the edges $\mathbf{E}$ are all undirected.  

 To build the node vectors, each cell at column $s_j \in \mathbf{S}$ of row $\mathbf{x}_i$ becomes a node with associated representation vector $\mathbf{v}=[j, \mathbf{X}_D(i,j)] \in \mathbb{R}^2$. The reason we encode the feature indices is to make the output graph permutation invariant. In other words, any change in the view or node order does not change the graph representation. Moreover, this makes the mapping between table row $\mathbf{x}_i$ and output graph $\textbf{G}_i$ bidirectional within $\mathbf{S}$ ($\mathbf{x}_i$ can be reconstructed from $\textbf{G}_i$ regardless of view into the graph). Our integration of feature indices in the graph is analogous to the design of positional encoding in transformer architecture \cite{vaswani2017attention} and its efficacy is shown in \cref{sec:ablation}. 

\noindent \textbf{Node pruning for CNV data} -- By default, all constructed graphs have the same edge structure, enforcing the same inductive bias through message passing when being modeled with GNNs. Extra domain knowledge could be leveraged to further strengthen this inductive bias. As in the case of CNV data, genes with the value of zero represent no variants in copy number as compared with the reference genes; therefore, they are usually not important for cancer subtyping. We drop all graph nodes with corresponding cell values equal to zero. This simplifies the graphs, diversifies the edge structure, and leads to better performance (see \cref{sec:ablation}).

Our X2Graph method is summarized in \cref{alg:x2graph}.

\subsection{Graph Modeling}
\label{sec:graph_model}
Given a graph $G_i$, we first project its node vectors $\mathbf{V}$ to an embedding space $\mathbf{P}$ of dimension $d$. Specifically, the first element (feature index) of $\mathbf{V}$ is transformed with an embedding layer $\mathcal{P}_0$ and the second element (tabular cell value) is transformed with a linear layer $\mathcal{P}_1$, each with dimension $d/2$:
\begin{equation}
    P = [\mathcal{P}_0(\mathbf{V}_{0}), \mathcal{P}_1(\mathbf{V}_{1})]_{\theta_0}
\end{equation}
where $[,]$ is the concatenation operation and $\theta_0$ is the learnable parameters of $\mathcal{P}_{0,1}$. Next, any network passing method can be applied to model the graphs. The exact architecture details are determined through parameter tuning (\cref{tab:arch}). Standard Cross Entropy loss is applied for our cancer subtyping problem.

\subsection{Multigraph Fusion}
\label{sec:graph_fusion}
As there are 3 separate KBs for CNV and RNA data, we adopt a late fusion approach to combine the predictions from each model trained for each KB. Specifically, our fusion model is a single matrix $W \in \mathbb{R}^{3\times 4}$ that produces a weighted sum over outputs $o^{(i)} \in \mathbb{R}^4\quad i=1,2,3$ of the $i$-th component models:
\begin{equation}
\bar{o}_j = \sum_{i} w_{ij}*o^{(i)}_j
\end{equation}
$W$ is trainable using gradient descent. Our fusion method is inspired by \cite{amer} but for the purpose of fusing multigraph models instead of multimodal fusion.




\section{EXPERIMENTS AND DISCUSSION}
\subsection{Dataset Settings and Evaluation Metrics}
We train and evaluate our models on the CNV, RNA and Clinical datasets described in \cref{sec:brca}. 
We perform 10-fold cross-validation, and the performance metrics are averaged across all folds. To address the unbalanced issue, we oversample the training data to match the size of the majority class. Validation and test sets are not oversampled. This is applied to X2Graph and all baselines for fair comparison. We report standard accuracy (micro), Area Under Curve (macro AUC), F1 score (macro) and Cohen Kappa ($\mathcal{K}_C$). Among these, AUC and F1 scores are measured for each class and averaged at the end, therefore accounting for class imbalance in the test set. $\mathcal{K}_C$ is a statistical metric widely used to measure the agreement between the predicted and ground truth labels, taking into account the agreement occurring by chance. All metrics have the upper threshold value of 1, and the higher it is, the better.     

\subsection{Training Procedure}
We trained X2Graph models using the PyTorch library. For CNV and RNA data, each model was trained for up to 200 epochs, while Clinical models were trained for 400 epochs, with early stopping applied when validation performance ceased to improve. We employed the Adam optimizer with an initial learning rate of $10^{-3}$ for CNV and RNA, and $10^{-4}$ for Clinical data, utilizing a cosine annealing schedule. The batch size was set to 64 for CNV and RNA, and for each training input graph, we randomly selected 6400 nodes and 3200 edges for data augmentation. Conversely, for Clinical data, the batch size was set to 16, and no random selection of nodes or edges was conducted. A hyperparameter search was performed with up to 50 random search iterations for X2Graph and other baselines. On a standard Intel Xeon 2.3GHz CPU server with 128GB RAM and a 48GB Quadro RTX 8000 GPU, it takes approximately 3 hours to complete a 10-fold cross-validation for an X2Graph experiment. 

\begin{figure*}[t]
    \centering
    \begin{tabular}{cc}
      \includegraphics[width=0.45\linewidth]{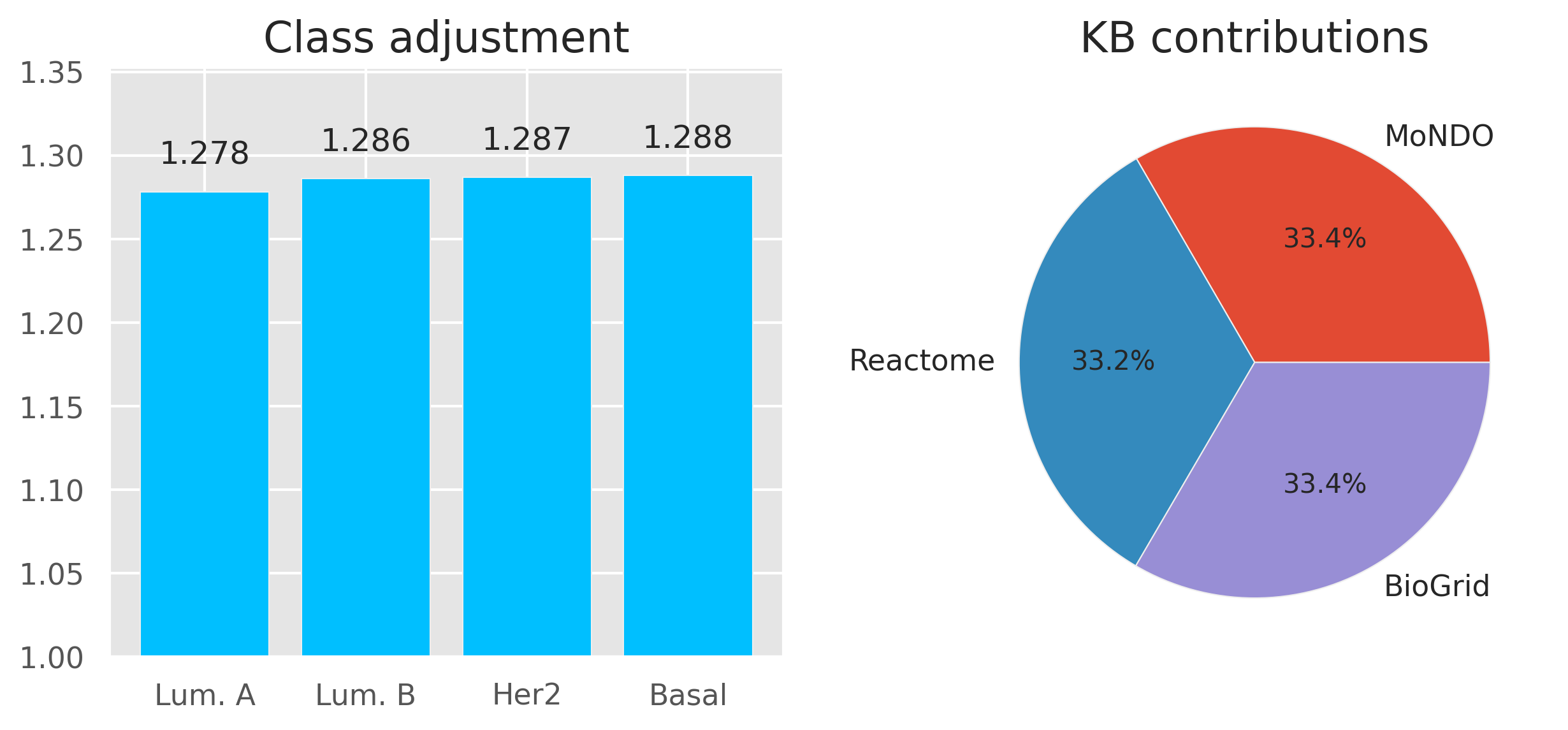}   &  
      \includegraphics[width=0.45\linewidth]{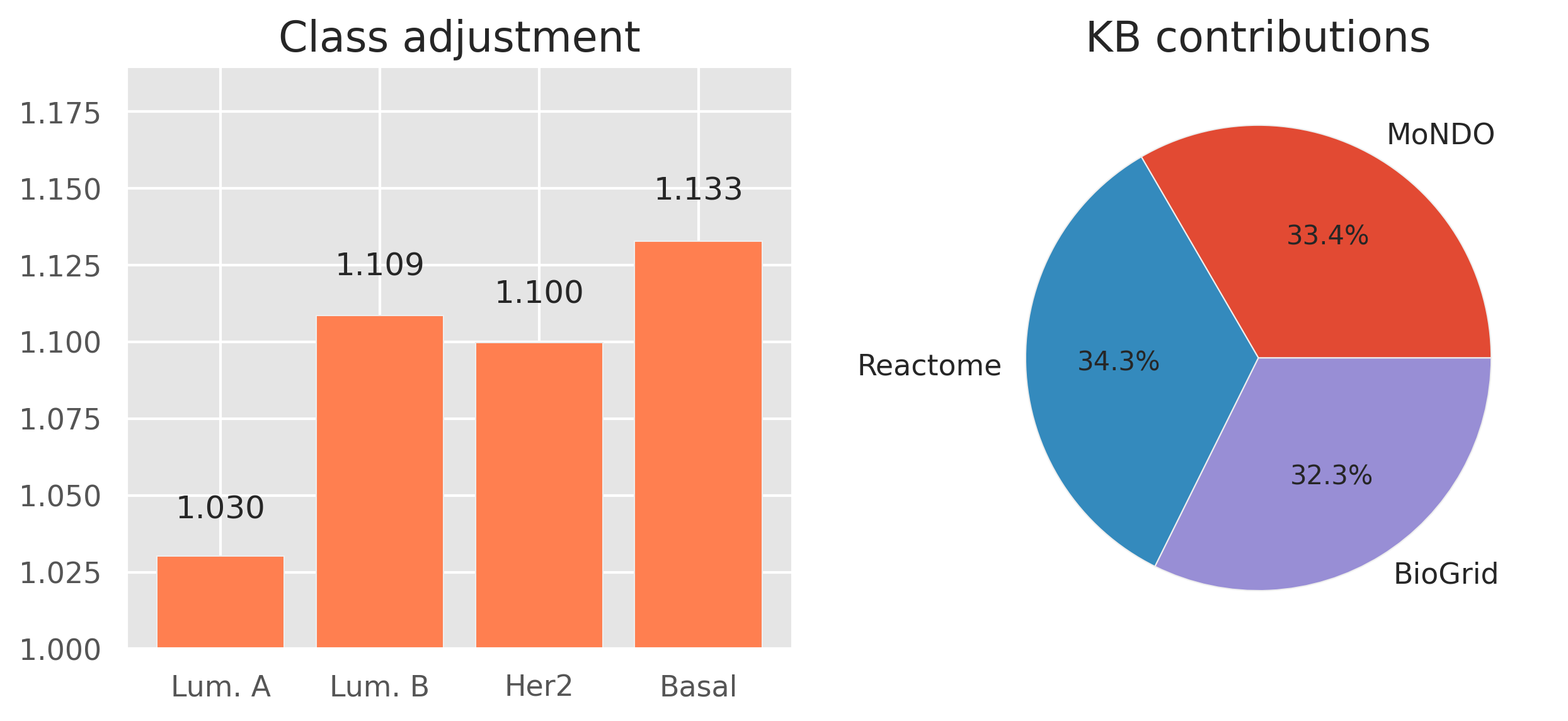} \\
      (a) CNV & (b) RNA
    \end{tabular}
    \caption{Contributions of each classes and KB in multigraph fusion for (a) CNV and (b) RNA, averaged across 10-fold cross validation.}
    \label{fig:fusion}
\end{figure*}
\subsection{Baseline Comparison}
We compare X2Graph with 2 deep learning -- MLP \cite{amer,wang2021lung,liu2022hybrid} and TabPFN \cite{hollmann2022tabpfn}, and 3 classical baselines -- XGBoost \cite{chen2016xgboost}, LASSO \cite{lasso} and RIDGE \cite{marquardt1975ridge}. MLP is benchmarked for its popularity in the biomedical domain, while the rest is selected as they are designed to work on small/medium tabular datasets. 

The results for CNV, RNA and Clinical datasets are shown in \cref{tab:cnv,tab:rna,tab:cli} respectively. While MLP outperforms LASSO and RIDGE on Clinical, it suffers a setback on datasets with small samples-to-features ratio such as CNV and RNA. TabPFN achieves better performance than MLP on several metrics -- \eg AUC on CNV and RNA, $\mathcal{K}_C$ on CNV -- but worse in others, especially on Clinical data. XGBoost, after properly tuned, has the highest performance among baselines. This is consistent with previous findings \cite{grinsztajn2022tree}. Finally, our proposed X2Graph outperforms all other models across various metrics, except for the F1 score on the Clinical datasets, where its performance is second to XGBoost on F1 but higher on accuracy, AUC and $\mathcal{K}_C$. On the CNV and RNA datasets, X2Graph models for individual KBs consistently outperform the MLP. This demonstrates that the strong inductive bias introduced by the KBs is the key factor in reducing overfitting and, consequently, enhancing performance. The Precision-Recall (PR) curves and Average Precision (AP) scores in \cref{fig:pr} also depict X2Graph's superiority versus other methods.

\subsection{Interpretability}
\label{sec:explain}
\noindent \textbf{Feature importance}. We employ GNNExplainer \cite{ying2019gnnexplainer} to identify the columns in each BioTD that contribute most to our X2Graph predictions. For simplicity, we experiment with fold 0 model (out of 10-fold cross-validation). For each graph sample in the test set, GNNExplainer approximates a subgraph that gives the closest prediction as the full graph. The selected nodes in the subgraphs are mapped back to the corresponding BioTD columns, thanks to its feature index (\cref{sec:x2graph}). \cref{fig:xai}(a-b) shows the top 10 most important features for CNV and RNA, respectively (out of more than 20k original features available in the BioTD). We further query the selected features on UNIPROT \cite{uniprot2025} to find possible links with breast cancer\footnote{Query format: https://www.uniprot.org/citations?query=BREAST+
CANCER+AND+\{X\} where X is the gene of interest}. Several genes are evidenced in literature for strong correlation with the disease; for example, TAZ is a breast cancer oncogene \cite{TAZ}, G6PD is highly expressed in breast cancer tissues \cite{G6PD}, and ADAM10 has recently been identified as a mediator for invasion and migration of breast cancer cells \cite{ADAM10}. The top 5 features for Clinical are also shown in \cref{fig:xai}(c). The most contributing features are `ER/PR status by IHC' which are clinically important markers for cancer treatment \cite{allison2020estrogen}.

We also compare the fraction of top genes with links to breast cancer for X2Graph and other two popular baselines - XGBoost and MLP in \cref{fig:xai}(d). For XGBoost, we identify the genes used most often for splitting across all decision trees \cite{chen2016xgboost}. For MLP, we employ Integrated Gradient \cite{integratedgradient} as the interpretability method. \cref{fig:xai}(d) indicates that X2Graph relies on more genes with evidenced links to breast cancer than other baselines (Clinical results are not shown as all features have a clear correlation with cancer).

\noindent \textbf{Multigraph fusion}. We visualize the contributions of each KB and class weight readjustment in the multigraph fusion models (\cref{sec:graph_fusion}) for CNV and RNA in \cref{fig:fusion}. All 3 KBs contribute almost equally to the predictions of the fusion models. For CNV, the class weights are balanced, while in the case of RNA, the weight of the Luminal A class is adjusted slightly lower than that of others. This indicates a slight bias of RNA models for individual KBs towards Luminal A, which is corrected in the fusion model.  
\squeezesmall
\subsection{Ablation}
\label{sec:ablation}
\vspace{-4mm}

\begin{table}[thpb]
\caption{Ablation of node pruning and gene ID indexing on X2Graph performance on CNV data with MoNDO KB.}
\label{tab:ablation}
\centering
\begin{tabular}{ll|cc}
\toprule
\multicolumn{2}{l}{\multirow{2}{*}{AUC / F1}} & \multicolumn{2}{c}{Node pruning} \\
\multicolumn{2}{l}{}                                                                          & \cmark               & \xmark              \\
\midrule
\multirow{2}{*}{\begin{tabular}[c]{@{}l@{}}ID\\ indexing\end{tabular}}          
    & \cmark       & \textbf{0.8892} / \textbf{0.6561}               & 0.8771 / 0.6460              \\
    & \xmark          & 0.7068 / 0.3783               & 0.7363 / 0.3847         \\
\bottomrule
\end{tabular}
\end{table}
We study the contributions of each X2Graph component to its overall performance. \cref{tab:ablation} shows the AUC and F1 scores of the CNV X2Graph model using MoNDO as the KB. We ablate the inclusion of gene ID in the node vectors and pruning of the diploid nodes (see \cref{sec:x2graph}). Including gene ID is essential for X2Graph as it improves AUC by at least 15\% and F1 score by at least 25\% regardless of node pruning status. We argue that the integration of gene ID provides the context for the model to encode its corresponding gene values, especially when the position of nodes in a graph changes due to node pruning (by design) or node dropping (as part of training augmentation). On the other hand, node pruning slightly improves the AUC/F1 performance by 1\% if gene ID is included. We attribute this behavior to the stronger inductive bias brought by the pruning of diploid genes. However, when ID indexing is turned off, it is better to also turn off node pruning so that the nodes appear in the same order across graphs, which benefits learning in the no-context scenario. 

To evaluate the efficacy of multi-graph fusion, we experiment with a modified X2Graph version where a single graph model with each edge being a multidimensional vector corresponding to the number of KBs. This multi-edge X2Graph version is considered as early fusion as opposed to the late fusion in our multi-graph approach. We empirically find that the multi-edge model underperforms the multi-graph, with AUC=0.8792 and F1=0.6137 on CNV data (\cf \cref{tab:cnv}). We conclude that, given the limited data, late fusion, as in our multi-graph, is a better strategy for enforcing the inductive bias from multiple KBs for X2Graph modeling. 

\section{LIMITATIONS}
X2Graph relies on an external knowledge source to construct graphs from tabular data, therefore is dependent on the quality of the KBs. The complexity of establishing a full KB for $n$ features is $O(n^2)$, which is challenging especially on domains whose feature relations are less known. Although we have demonstrated that X2Graph works well even on incomplete KBs (as for CNV, RNA and also in handcrafted-KB for Clinical data), our assumption is that the most crucial relations between features are already defined in these KBs. The capability of exploring novel feature relations is also beyond the scope of X2Graph.  

\section{CONCLUSION}
We propose X2Graph as an efficient modeling method for small tabular data. X2Graph introduces inductive bias to model training through predefined connections between table columns, effectively mitigating the overfitting problem when training with limited data. Our results demonstrate the efficacy of X2Graph on three real-world biological datasets where prior knowledge about column connections is limited. Achieving state-of-the-art performance, X2Graph presents a promising deep learning alternative to traditional statistical and tree-based methods that have dominated tabular data analysis. This advancement has the potential to significantly impact medicine and biology by enhancing the precision of cancer subtyping and other diagnostic processes. Future work includes extending X2Graph to other tabular data domains and integrating it into multimodal learning frameworks, where samples from various modalities can be converted into graph representations.    

\bibliographystyle{IEEEtran}
\bibliography{main}

\end{document}